\documentclass[journal,twoside,web]{ieeecolor}
\usepackage{generic}
\usepackage{cite}
\usepackage{amsmath,amssymb,amsfonts}
\usepackage{algorithmic}
\usepackage{graphicx}
\usepackage{algorithm,algorithmic}
\usepackage{hyperref}
\hypersetup{hidelinks=true}
\usepackage{textcomp}

\usepackage{tabularx}
\usepackage{array}
\usepackage{booktabs}
\usepackage{float}
\usepackage{cuted}
\usepackage[table]{xcolor}
\usepackage{soul}
\usepackage{dblfloatfix} 
\usepackage{comment}
\usepackage{capt-of}

\definecolor{lightred}{RGB}{255,200,200}
\definecolor{lightpink}{RGB}{255,225,235}

\def\BibTeX{{\rm B\kern-.05em{\sc i\kern-.025em b}\kern-.08em
    T\kern-.1667em\lower.7ex\hbox{E}\kern-.125emX}}
\providecommand{\refname}{References} 

\begin{document}

\title{\textsc{Watch}, \textsc{Learn}, \textsc{Assist}: Unlocking Autonomous Prosthetic Manipulation via Imitation Learning}

\author{Martina Columbaro\textsuperscript{1,3},
        Federico Vasile\textsuperscript{1},
        Nicolò Boccardo\textsuperscript{2},
        Lorenzo Natale\textsuperscript{1} 
\thanks{\textsuperscript{1} Humanoid Sensing and Perception, Istituto Italiano di Tecnologia, Genoa, Italy (email: \texttt{\href{mailto:name.surname@iit.it}{name.surname@iit.it}} )}
\thanks{\textsuperscript{2} Rehab Technologies Lab, Istituto Italiano di Tecnologia, Genoa, Italy (email: \texttt{\href{mailto:name.surname@iit.it}{name.surname@iit.it}} )}
\thanks{\textsuperscript{3} Dipartimento di Informatica, Bioingegneria, Robotica e Ingegneria dei Sistemi (DIBRIS), University of Genova, Genoa, Italy.}
\thanks{This work has been submitted to the IEEE for possible publication. Copyright may be transferred without notice, after which this version may no longer be accessible.}
}

\maketitle

\begin{strip}
    \vspace{-2.5cm}
    \centering
    \includegraphics[width=1\textwidth]{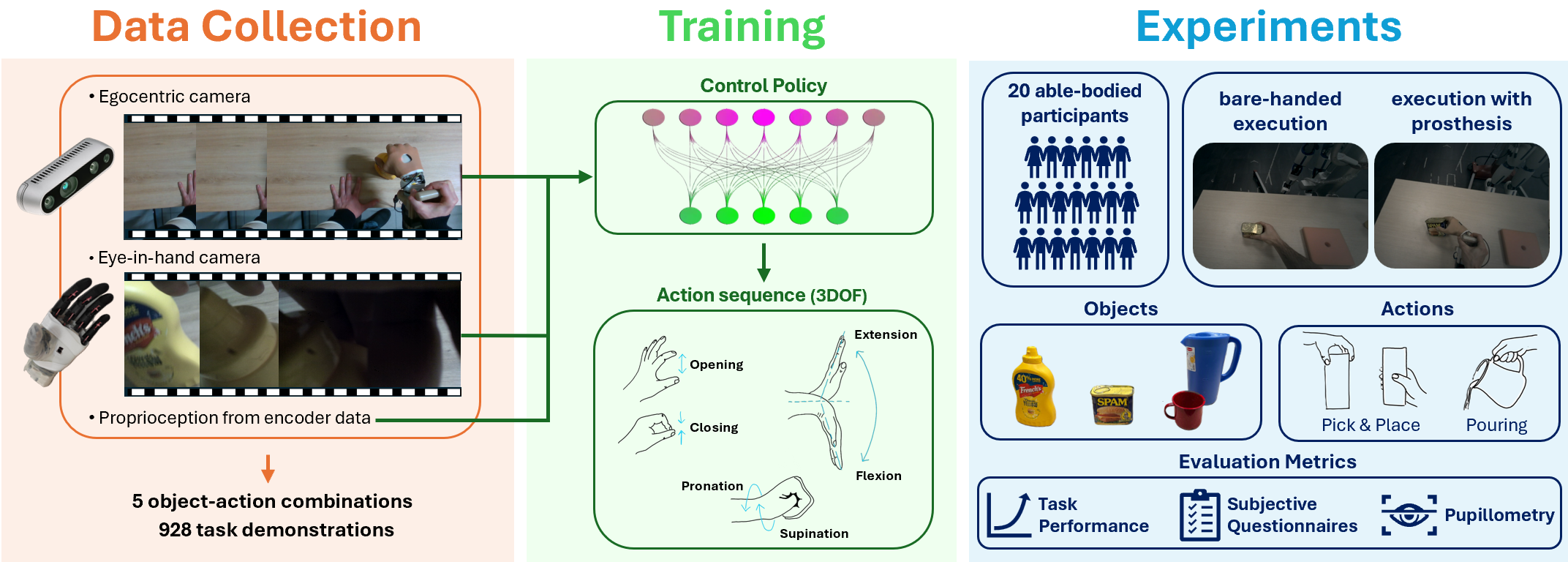}
    \captionof{figure}{Overview of the proposed framework. Manipulation demonstrations are collected using egocentric vision, eye-in-hand vision and prosthetic proprioception. These multimodal data are used to train an imitation-learning policy that predicts 3-DoF prosthetic commands. 
    The policy is evaluated with able-bodied participants wearing the prosthesis. Its performance is compared with bare-handed task execution using task-performance metrics, subjective questionnaires and physiological measures.}
    \label{fig:main_fig}
\end{strip}

\begin{abstract}
Object grasping and manipulation with a prosthetic hand remain challenging for individuals with transradial limb loss, particularly when multiple prosthetic degrees of freedom must be coordinated. We present a framework for autonomous control of a prosthetic hand during grasping and manipulation tasks. Our approach leverages imitation learning to train a control policy that generates motor commands for the prosthesis from egocentric images, eye-in-hand images and prosthetic proprioception. 
The user guides the hand through the workspace, while the policy autonomously controls the degrees of freedom to complete the task. 
To train the policy, we collect a dataset of manipulation demonstrations using a teleoperation-inspired, two-operator setup. The dataset covers five object-action combinations with different grasping and manipulation requirements. We evaluate the policy with 20 able-bodied participants, who perform the same tasks both bare-handed and while wearing the prosthesis. Results show that the system achieves high task-completion rates across the evaluated tasks, with rates between 78.5\% and 97\%, and progressive success rate above 92\%. Subjective questionnaire responses further indicate that the system is perceived as easy to learn and manageable to use. Our results mark a step forward, providing evidence that multimodal control policies can support autonomous prosthetic control and restore complex coordination capabilities in prosthetic users.
\end{abstract}


\section{Introduction}
\label{sec:introduction}
\IEEEPARstart{I}{n} recent years, significant progress has been made in the control of upper-limb prostheses, particularly for object grasping. Vision-based methods and semi-autonomous control strategies have demonstrated that external sensory information can effectively complement myoelectric inputs, reducing users' cognitive effort and improving task performance~\cite{vasile2025continuous,10472622,xu2025powered,stracquadanio2025bring,11184238,karrenbach2022improving}. However, while these approaches have achieved promising results in \textit{short-horizon} tasks such as object grasping, \textit{multi-step} object manipulation has received considerably less attention. This is a critical limitation, since many activities of daily living require the prosthetic hand to perform a sequence of interactions with the same object to complete a task. Today, the increasing dexterity of modern prosthetic devices~\cite{laffranchi2020hannes,Yang2025Lightweight}, together with recent advances in learning-based control~\cite{chi2024diffusionpolicyvisuomotorpolicy}, creates new opportunities to address this challenge.

Moving from grasping to complex manipulation introduces substantial perceptual and control challenges. Vision-based prosthetic grasping systems have often relied on eye-in-hand cameras, which provide close-range observations of the object to grasp and support accurate inference of grasp intent~\cite{CIRELLI2026104669,10522893,robotics15010022,app16125736}. However, in complex manipulation tasks, where the prosthesis may hold an object and interact with the contralateral hand or the environment, the only hand-mounted camera becomes insufficient: fundamental elements of the interaction may remain occluded or outside the field of view during the task. This limitation calls for the introduction of an egocentric camera, which complements the local perspective by capturing the broader task context, including both hands, the manipulated objects and their interactions with the environment. 

Egocentric vision captures the environment from the user's perspective and therefore provides a natural view of how actions unfold during everyday activities. Large-scale datasets~\cite{damen2018scaling,grauman2022ego4d} have enabled substantial progress in recognizing ongoing activities, anticipating future actions and modeling hand-object interactions. Beyond assigning semantic labels to video segments, recent methods have investigated the prediction of future hand trajectories and object-interaction regions~\cite{Liu_2022_CVPR,10.1007/978-3-031-73202-7_4}. These developments demonstrate that egocentric observations contain information about both the current state of an interaction and its likely evolution. For prosthetic manipulation, this perspective is particularly valuable because it can capture the relative motion of the prosthesis and the contralateral hand, changes in object state and the progression of a multi-step task.

Despite these advances, most egocentric video datasets are designed primarily for visual understanding rather than direct robot control and lack information about interaction forces. This makes them useful for pre-training on visual tasks, but limits their applicability to learning low-level motor control policies. Imitation learning from teleoperated demonstrations offers a solution to this problem. By collecting synchronized observations and motor commands while the target device performs a task, imitation learning can optimize a policy that maps sensory observations directly to device-specific commands.

In this work, we combine egocentric vision, eye-in-hand vision and prosthetic proprioception, and formulate control as conditional action-sequence generation. In particular, we adopt a diffusion-based visuomotor policy~\cite{chi2024diffusionpolicyvisuomotorpolicy,song2021denoising}, which models the distribution of future prosthetic commands conditioned on a short history of multimodal observations. The user guides the prosthesis through the workspace, while the learned policy autonomously controls the wrist and finger degrees of freedom required to perform the task. Our contributions are as follows:

\begin{itemize}
    \item We propose a multimodal diffusion-policy model for prosthetic grasping and manipulation. The model combines egocentric vision, eye-in-hand vision and prosthetic proprioception to generate motor commands for a poly-articulated prosthetic hand (the Hannes prosthetic hand~\cite{laffranchi2020hannes}).
    
    \item We collect a dataset of coordinated manipulation demonstrations with the Hannes prosthesis, covering five object-action combinations and auxiliary recovery behaviors.

    \item We train a single multi-task policy across all object-action combinations. The policy predicts commands for the three degrees of freedom (DoFs): wrist flexion-extension, wrist pronation-supination and fingers opening-closing.

    \item We evaluate the proposed framework in a user study involving 20 able-bodied participants without upper-limb loss, using task-performance metrics, subjective questionnaires and pupillometry.
\end{itemize}

In Sec.~\ref{sec:related_work}, we review the state of the art on vision-based systems for prosthetics and egocentric learning in robotics. Sec.~\ref{sec:dataset} describes the data-collection procedure and the resulting dataset. The proposed multimodal diffusion-policy model is introduced in Sec.~\ref{sec:methods}. Sec.~\ref{sec:mainresults} and Sec.~\ref{sec:furtheranalisys} present the experimental evaluation and results. Finally, Sec.~\ref{sec:conclusions} summarizes the main conclusions and outlines directions for future work.

\section{Related Work}
\label{sec:related_work}
\begin{figure*}[t]
    \centering
    \includegraphics[width=1\linewidth]{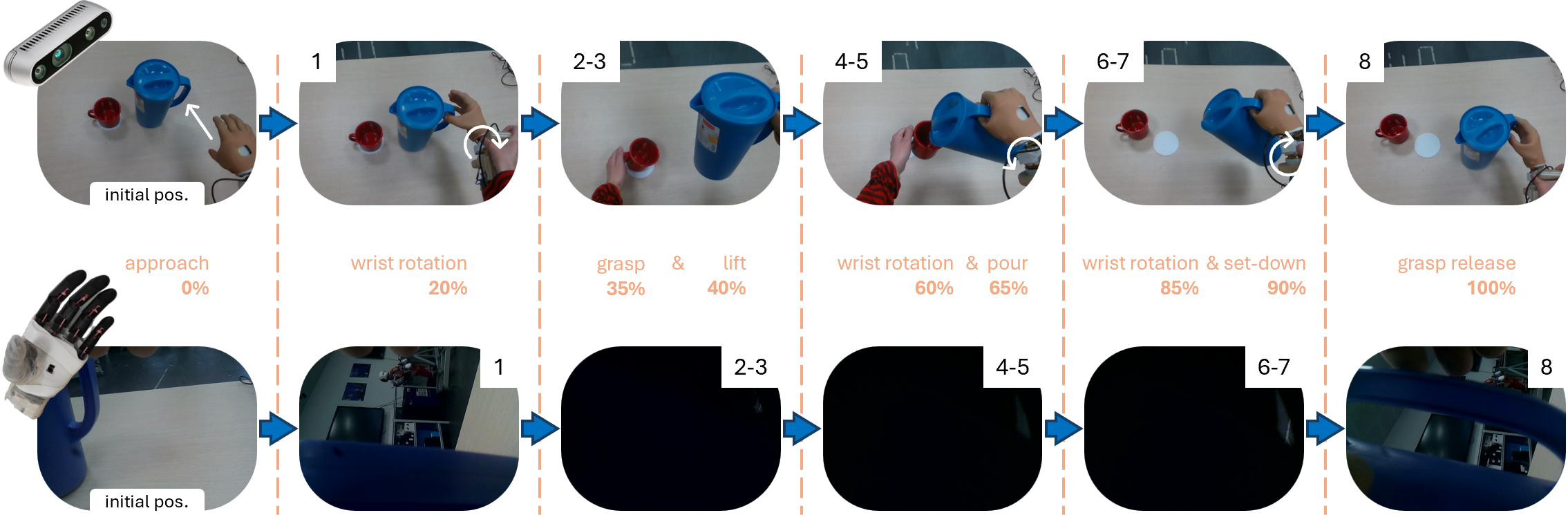}
    \caption{Example progression of the pouring task. The top row shows the egocentric view, while the bottom row shows the corresponding eye-in-hand view from the camera embedded in the prosthetic palm. Vertical dashed lines separate the intermediate steps used for the computation of the PSR. Percentages indicate the progressive percentage assigned to each step.}
    \label{fig:pouring_example}
    \vspace{-0.2cm}
\end{figure*}    

\subsection{Vision-based Prosthetic Control}
Commercially available upper-limb prostheses are commonly controlled through surface electromyography (sEMG), which translates residual-muscle activity into commands for the prosthetic joints. 
However, changes in electrode placement and limb position can degrade sEMG-based control performance~\cite{s24154840}. Moreover, conventional myoelectric controllers often operate prosthetic functions sequentially, requiring users to switch explicitly between them~\cite{farina2021higher}.
Consequently, additional sensing modalities have been investigated to provide contextual information and reduce the control burden placed on the user.
For example, inertial measurements can complement sEMG by capturing residual-limb motion and have been shown to improve both motion classification and real-time prosthesis control~\cite{krasoulis2017improved, volkmar2019improving,zaman2026voluntary}.
Alternatively, visual information provides a particularly informative source of context because it can capture the target object and its spatial relationship with the prosthesis before physical contact occurs~\cite{vasile2022,11114342,Karaali2025VisualLanguageModels,10.1145/3762657}. Furthermore, it can be combined with myoelectric information to infer grasp configurations or other control parameters, thereby allowing the prosthesis to automate selected components of the task while retaining user control over grasp initiation or execution~\cite{markovic2015sensor,starke2022semi}. Most of these systems implement shared or semi-autonomous control: vision determines a grasp type, preshapes the hand or assists wrist orientation, whereas the user remains responsible for triggering actions or controlling other phases of the task. Recent imitation-learning approaches have moved toward greater autonomy by relying primarily on hand-mounted cameras to perform grasp-and-release tasks~\cite{alessi2025hannesimitation, shi2026biosignals}. 

Overall, existing vision-based prosthetic controllers are predominantly semi-autonomous or are limited to short-horizon tasks. In contrast, we propose a fully autonomous control policy for both short and long horizon tasks, including object pick-and-place and pouring into a mug. 

\subsection{Egocentric Learning in Robotics}

Imitation learning provides a direct approach to acquiring manipulation skills by training a policy on demonstrations consisting of synchronized observations and actions. Its effectiveness, however, depends strongly on the quantity, diversity, and quality of the demonstrations available for the target robot. Conventional teleoperation systems can provide accurately aligned robot observations and commands, but collecting such data is labor-intensive. A complementary strategy is to first learn broadly useful representations or behavioral priors from data that are easier to acquire, and subsequently adapt them using a smaller set of demonstrations aligned with the target robot. Egocentric videos of humans manipulating objects are particularly attractive for this purpose. They can be collected at scale in everyday environments and provide a task-oriented view of hands, objects and their interactions. Early methods showed that visual representations pretrained on large human-video or human-centric image datasets can improve data efficiency across downstream robot-manipulation tasks~\cite{nair2022r3m,Radosavovic2022}. More recent methods have investigated how egocentric human demonstrations can contribute more directly to policy learning by combining human and robot demonstrations, defining compatible human-robot action spaces, or introducing explicit alignment procedures~\cite{pmlr-v229-wang23a,kareer2025egomimic,yang2025egovla}.

Following these developments in egocentric robot learning, we combine a vision encoder pretrained on egocentric human data with eye-in-hand vision and prosthetic proprioception to predict motor commands. To the best of our knowledge, the joint use of egocentric and eye-in-hand observations within an imitation-learning policy for multi-step prosthetic manipulation has not previously been investigated.

\section{Dataset}
\label{sec:dataset}
\subsection{Data Collection Setup}

\noindent\textbf{Object Set.}
We select objects with different geometries, sizes and manipulation requirements from the YCB object set~\cite{calli2015benchmarking}. The selected objects include a mustard bottle, a potted meat can, a pitcher and a mug. These objects represent common household items while requiring distinct hand configurations and interaction strategies.

\noindent\textbf{Manipulation Tasks.}
The dataset comprises both grasping and manipulation tasks. For the mustard bottle and the potted meat can, pick-and-place demonstrations are collected using two different grasp configurations: a top grasp and a side grasp. These configurations are used to assess whether the control policy can infer user intent from visual observations. Specifically, the policy should orient the wrist according to whether the user approaches the object from the top or the side. 
The dataset also includes a bimanual manipulation task in which the prosthesis grasps a pitcher and pours its contents into a mug. This task requires coordinated interaction between the intact hand and the prosthesis: the former grasps and holds the mug, while the latter performs a stable grasp on the pitcher and executes the pouring motion (Fig.~\ref{fig:pouring_example}).

\begin{figure}[H] 
    \centering 
    \includegraphics[width=1\linewidth]{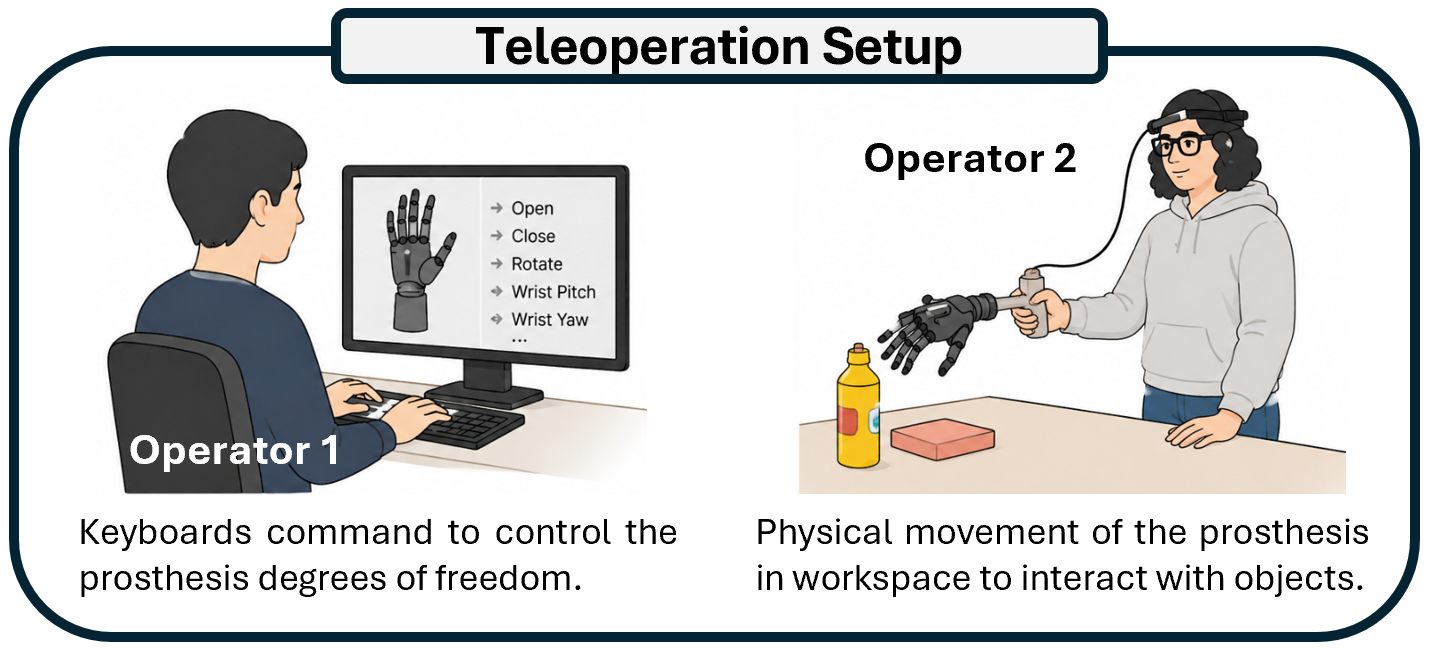} 
    \caption{Two-operator setup used for demonstration collection. 
             } 
    \label{fig:datacollection_operators}
\end{figure}

\noindent\textbf{Data collection.} Demonstrations are collected using a teleoperation-inspired, two-operator setup (Fig.~\ref{fig:datacollection_operators}). 
The first operator controls the prosthesis DoFs via keyboard commands, actuating the joints required for grasping and manipulation. Meanwhile, the second operator guides the Hannes prosthetic hand through the workspace using an able-bodied adapter. For instance, during a pick-and-place task, the second operator guides the hand toward the object, while the first operator controls the DoFs to rotate the wrist and close the fingers for grasping. Then, the second operator transports the object to the target location, where the first operator commands fingers opening to release it. Thus, task execution is achieved through coordinated control of the prosthesis by the two operators. 

We remark that our objective is to learn an autonomous policy that acts in compliance with the user’s movements. Accordingly, during data collection, the second operator is responsible solely for moving the prosthesis through the workspace. Assigning this operator also joint-level control would introduce a discrepancy between data collection and the policy’s intended behavior at deployment. Therefore, the first operator commands the prosthesis joints, closely emulating the control behavior that we want the policy to learn.

During each demonstration, egocentric and eye-in-hand visual observations are recorded, together with prosthetic proprioception and the commanded actions. Data acquisition and synchronization across sensors and control streams are handled using the YARP middleware~\cite{mettayarp}. We highlight that both views are essential for the considered tasks. Specifically, the head-mounted camera provides global cues of the workspace, whereas the eye-in-hand camera captures local information about the object and the prosthesis-object interactions.
 
\subsection{Dataset Statistics}

Table~\ref{tab:demo_durations} summarizes the number of demonstrations collected for each task and their average duration. Each task is defined by a specific object–action pair. PP-Top and PP-Side denote pick-and-place actions performed using top-grasp and side-grasp configurations, respectively.

\begin{table}[H]
\centering
\caption{Demonstration statistics for each task. Duration is reported as mean $\pm$ standard deviation in seconds.}
\label{tab:demo_durations}
\resizebox{\columnwidth}{!}{%
\begin{tabular}{cllcc}
\toprule
Task ID & Object & Action & \# Demos & Duration [s] \\
\midrule
1 & Mustard Bottle & PP-Top & 201 & $13.46 \pm 1.15$ \\
2 & Mustard Bottle & PP-Side & 204 & $17.32 \pm 2.36$ \\
3 & Potted Meat Can & PP-Top & 160 & $12.40 \pm 1.32$ \\
4 & Potted Meat Can & PP-Side & 160 & $15.94 \pm 1.80$ \\
5 & Pitcher & Pouring & 203 & $24.22 \pm 2.92$ \\
\bottomrule
\end{tabular}%
}
\end{table}

The dataset comprises 928 nominal task demonstrations distributed across five object–action combinations. In addition, we collected 48 auxiliary demonstrations of recovery behaviors. Because these demonstrations are not associated with a specific object–action pair, they are not included in Table~\ref{tab:demo_durations}. Their addition follows common practice in robotics~\cite{argall2009survey}: they provide examples of corrective motions that return the prosthesis to a valid configuration, or maintain it in the home configuration when it deviates from the nominal demonstrations. Note that even though we collect these demonstrations offline rather than through online dataset aggregation~\cite{ross2011dagger}, the underlying principle remains applicable: they expose the policy to corrective actions for recovering from states outside the distribution of nominal demonstrations.

The dataset was collected by two demonstrators over five acquisition sessions.
The number of demonstrations was adjusted iteratively across sessions, with additional samples collected for tasks that showed more frequent execution errors during intermediate policy evaluations. 
The use of multiple demonstrators introduces differences in arm-level positioning, object approach, execution speed and coordination between the intact hand and the prosthesis. To introduce variability in both human execution and visual appearance, the sessions were recorded using different workspace backgrounds and scene configurations. This variation changes the visual context observed by the egocentric and eye-in-hand cameras while preserving the structure of the manipulation tasks.  Overall, the dataset contains demonstrations spanning different objects, grasp configurations, operators and visual conditions.

\section{Control Policy}
\label{sec:methods}
\begin{figure}[H]
    \centering
    \includegraphics[width=1\linewidth]{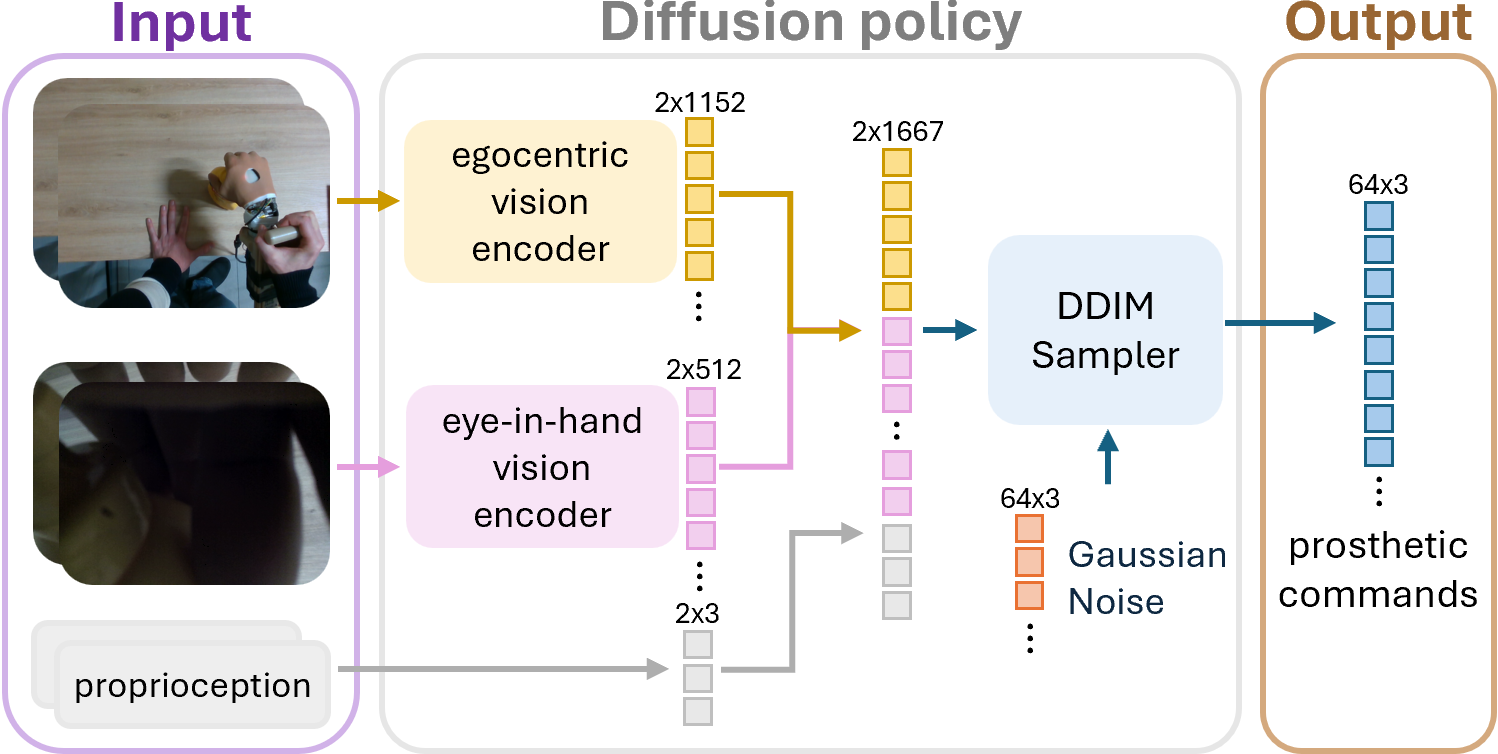}
    \caption{Overview of the proposed diffusion-based policy.  A two-step history of egocentric and eye-in-hand images is encoded and then concatenated with the prosthetic proprioception to condition a DDIM sampler. This one generates a future sequence of 3-DoF prosthetic commands.}
    \label{fig:method_overview}
\end{figure}

The proposed framework learns a visuomotor policy for manipulation tasks with the Hannes prosthetic hand. As illustrated in Fig.~\ref{fig:method_overview}, the policy receives three synchronized input streams: the image from the head-mounted camera, the eye-in-hand image from the camera embedded into the prosthesis palm and the proprioceptive state of the prosthesis. From these observations, the policy predicts motor commands for the three DoFs: wrist flexion-extension (FE), wrist pronation-supination (PS) and fingers opening-closing (OC). A single multi-task policy is trained across all object-action combinations.

The two visual streams are processed by separate vision encoders. The eye-in-hand observations are highly specific to our setup, as they primarily contain close-range views of the object and local hand-object interactions. Given this domain specificity, we use a ResNet-18~\cite{he2016deep} trained from scratch on our demonstrations, allowing its visual representation to adapt directly to the task domain. For the egocentric stream, we leverage a vision encoder pretrained on large-scale egocentric human-manipulation data. Specifically, EgoVLA~\cite{yang2025egovla} trains a vision-language-action model to predict future wrist and hand configurations from a history of egocentric images, the current hand state and a language instruction. We extract the SigLIP vision encoder~\cite{zhai2023sigmoid} from the pretrained EgoVLA model and integrate it into our control policy. This initialization provides the policy with visual features learned from diverse egocentric manipulation scenarios before adaptation to the target tasks.

The overall control policy follows a diffusion-based architecture~\cite{chi2024diffusionpolicyvisuomotorpolicy}. At each control cycle, the vision encoders map the two visual streams to latent representations, which are concatenated with the prosthetic proprioceptive state to form a multimodal observation. Conditioned on a short history of these observations, a DDIM sampler~\cite{song2021denoising} iteratively transforms a sequence of Gaussian noise into a sequence of future prosthetic commands. In our experiments, the observation horizon is set to two time steps, and the prediction horizon is set to 64 time steps. The policy operates in a receding-horizon, closed-loop manner. At each iteration, three commands of the predicted sequence are sent to the prosthesis with a time interval of 0.09~s between consecutive commands, allowing the prosthesis sufficient time for their execution. 
This process is repeated until the task is completed.

The policy is trained using the dataset described in Sec.~\ref{sec:dataset}. Each demonstration is divided into training samples comprising a short observation history and the corresponding future action sequence. The proprioceptive states and action commands are normalized to the interval $[-1,1]$ using the admissible ranges of the controlled DoFs. We train the policy for 22 epochs using the AdamW~\cite{loshchilov2019decoupled} optimizer with a learning rate of $10^{-4}$. The pretrained egocentric vision encoder is initially kept frozen and is fine-tuned during the final 2 epochs using a learning rate of $10^{-6}$.

The eye-in-hand images are resized to $256 \times 192$ pixels before being passed to the ResNet-18 encoder. For the egocentric stream, we adhere to the preprocessing procedure adopted by EgoVLA~\cite{yang2025egovla}: during training, the images are resized and randomly cropped to $384 \times 384$ pixels, whereas a center crop is used during inference. The diffusion model is trained with a noise schedule of 100 diffusion steps, while 20 denoising steps are used at inference time.

\section{Results}
\label{sec:mainresults}
\subsection{Experimental Setup}

We conducted a user study on 20 able-bodied participants without upper-limb loss to evaluate the proposed framework for prosthetic manipulation. 
All participants provided informed consent prior to the study. The experimental protocol was conducted in accordance with the Declaration of Helsinki and approved by the Ethical Committee of Liguria Region, Italy (approval n.~363/2022, DB ID 12494).  
Each participant performed the experimental tasks under two conditions: 
\begin{itemize}
    \item \textit{bare hands}: the user performs the tasks using their hands;
    \item \textit{prosthesis}: the user wears an able-bodied adapter with the Hannes hand attached to it. Note that the pouring task requires interaction with their contralateral bare hand. 
\end{itemize}
The \textit{bare hands} condition is included as a reference for comparing task performance and pupil responses with those observed during prosthesis use. During the \textit{prosthesis} condition, participants wear the egocentric camera and physically move the prosthesis within the workspace through the able-bodied adapter while the control policy commands the DoFs. The prosthetic hand embeds a camera in the palm, providing a local view of the target object and the approaching direction. Both the egocentric and eye-in-hand views are provided as input to the control policy. This configuration reproduces the setup used during data collection.

Participants wear Tobii eye-tracking glasses (Tobii AB, Stockholm, Sweden; \href{https://www.tobii.com/products/eye-trackers/wearables/tobii-pro-glasses-3}{Tobii Pro Glasses 3}) throughout the experiment. The device is used to record pupil-diameter variations during task execution as a physiological correlation with cognitive and physical demand~\cite{wel2018pupil}. Each participant is asked to perform the same object-action combination with the two conditions previously introduced (i.e., \textit{bare hands} and \textit{prosthesis}), allowing pupil responses under the \textit{prosthesis} condition to be compared with a reference. Because absolute pupil diameter varies considerably across individuals and may also be affected by environmental and physiological factors, the analysis is performed using participant-specific changes relative to a baseline rather than raw pupil-size values.
To this end, a separate pupil baseline was collected immediately before the \textit{bare hands} condition and immediately before the \textit{prosthesis} condition. Pupil measurements recorded during the experimental trials were expressed relative to the baseline acquired at the beginning of that condition. This procedure accounts for differences in initial pupil diameter between participants and for possible changes between the two experimental blocks, enabling a more reliable within-subject analysis. 

Before starting the prosthetic trials, each participant completed approximately 15 minutes of familiarization with the system. Similar familiarization or training phases are commonly included in prosthetic-control evaluation protocols~\cite{lyons2018upper, trout2025shared, shehata2018improving}. During this phase, participants learned how the prosthesis responded, practiced coordinating it with the intact hand and performed examples of the required manipulation tasks. Familiarization trials are not included in the analysis. During the experimental phase, each object-action combination was repeated ten times.

Our analysis combines complementary measures to provide a comprehensive evaluation of the proposed system. Specifically, we consider task-performance measures, subjective questionnaires and physiological measurements. Task performance is assessed through the binary success rate (SR), the progressive success rate (PSR) and task execution time. SR indicates the number of trials in which the task is fully accomplished, whereas PSR provides a more fine-grained measure by accounting for the last intermediate stage completed during unsuccessful trials. For the computation of PSR, each task is divided into a sequence of sub-tasks reflecting its expected progression. PP-Top trials are evaluated over 6 intermediate steps, while PP-Side and pouring (Fig.~\ref{fig:pouring_example}) trials are evaluated over 8 intermediate steps.
Execution time is used to evaluate the temporal efficiency of the interaction. 
Subjective workload is assessed through the NASA Task Load Index (NASA-TLX), perceived usability through the System Usability Scale (SUS) and additional aspects specific to the proposed system through an ad-hoc questionnaire.
Finally, variations in pupil diameter are analyzed as a physiological indicator of cognitive and physical effort.

\subsection{Success Rate}

\begin{figure}[]
    \centering
    \includegraphics[width=1\linewidth]{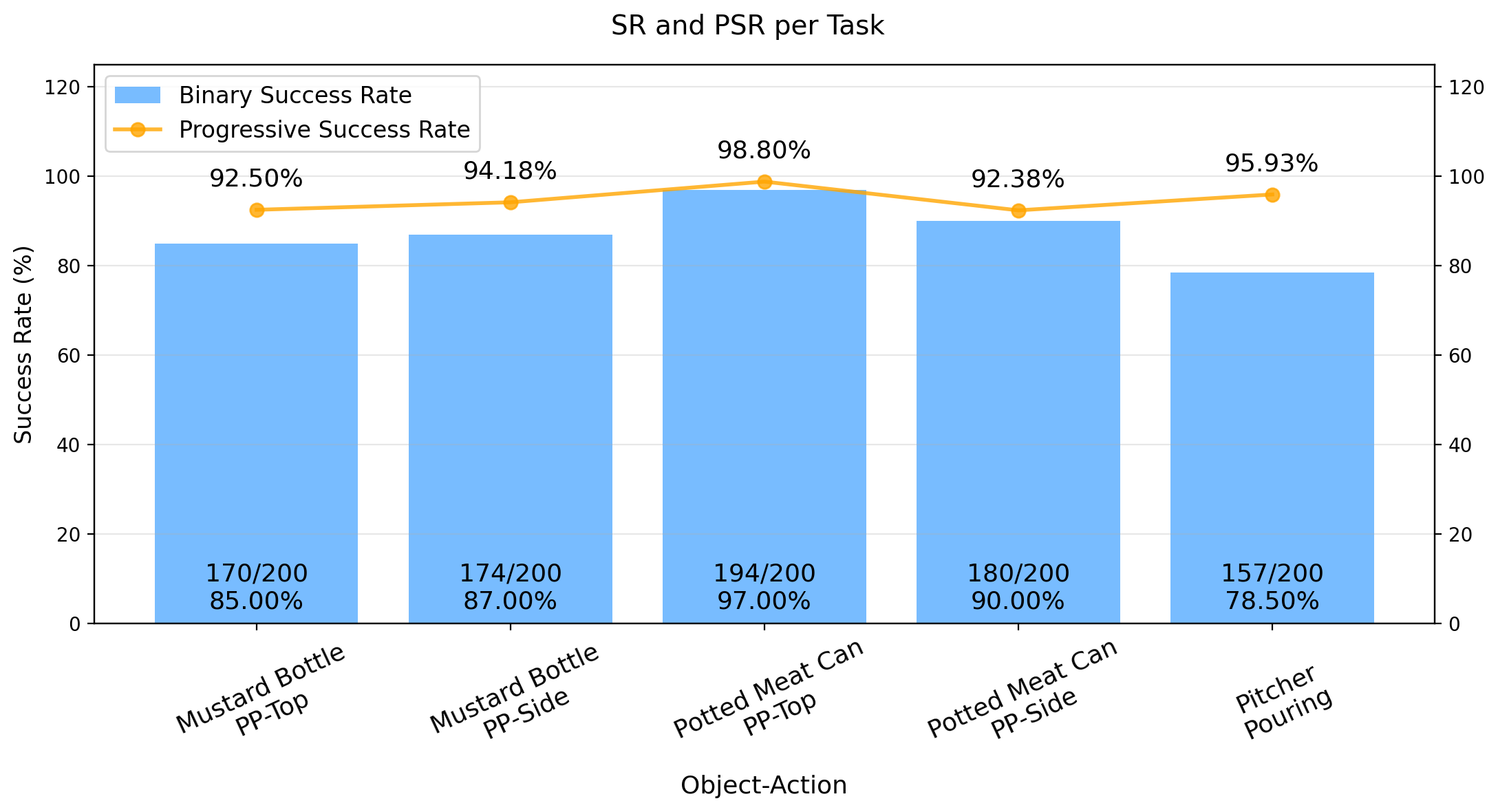}
    \caption{Binary success rate (SR) and progressive success rate (PSR) for each object-action combination during the \textit{prosthesis} condition. }
    \label{fig:plot_success_rate}
\end{figure}

Fig.~\ref{fig:plot_success_rate} reports the success rate (SR) and the progressive success rate (PSR) obtained during the \textit{prosthesis} condition for each object-action combination. Overall, participants achieved high completion rates across the evaluated tasks. The SR ranged from 78.5\% to 97.0\%, while the PSR ranged from 92.38\% to 98.80\%.

The highest SR was obtained for the potted meat can PP-Top task, for which participants successfully completed 194 out of 200 trials (97.0\%). This is reasonable, being the easiest task with the object with the easiest shape.
The pouring task performed with the pitcher was the most challenging in terms of complete task accomplishment, with an SR of 78.5\% (157/200 trials).

For all tasks, the PSR was higher than the corresponding binary SR. This indicates that, even when participants did not complete the entire task, they were generally able to execute it up to a certain stage before failing. This behavior was particularly evident for the pouring task, during which the SR was the lowest among the evaluated tasks, whereas the PSR remained high at 95.93\%. 

The progression of the pouring task is illustrated in Fig.~\ref{fig:pouring_example}. In many unsuccessful trials, participants completed the approach, wrist rotation, grasping, lifting and pouring phases, but then failed during the final set-down or release steps.
Therefore, failures in this task were generally associated with difficulties in the final stages of the manipulation rather than with an inability to initiate or substantially progress through the task. 

The relatively small difference between SR and PSR for the potted meat can PP-Top task further confirms that this was the most consistently completed task. In contrast, the larger gaps, observed for the mustard bottle and for the pouring task (Fig.~\ref{fig:plot_success_rate}), suggest that these interactions were more sensitive to prosthesis positioning and grasp stabilization. In these cases, participants often progressed through most of the task but failed during the final stages.
For the pouring task, this behavior was partly related to the actuation mechanism of the prosthetic hand. Since the four fingers are driven by a single motor, they extended simultaneously when the hand opens. As a result, if the pitcher was imperfectly positioned or not enough stabilized at release, the simultaneous extension could push the object and, in some cases, cause it to fall. 

\subsection{Execution Time}

\begin{figure}[H]
    \centering
    \includegraphics[width=1\linewidth]{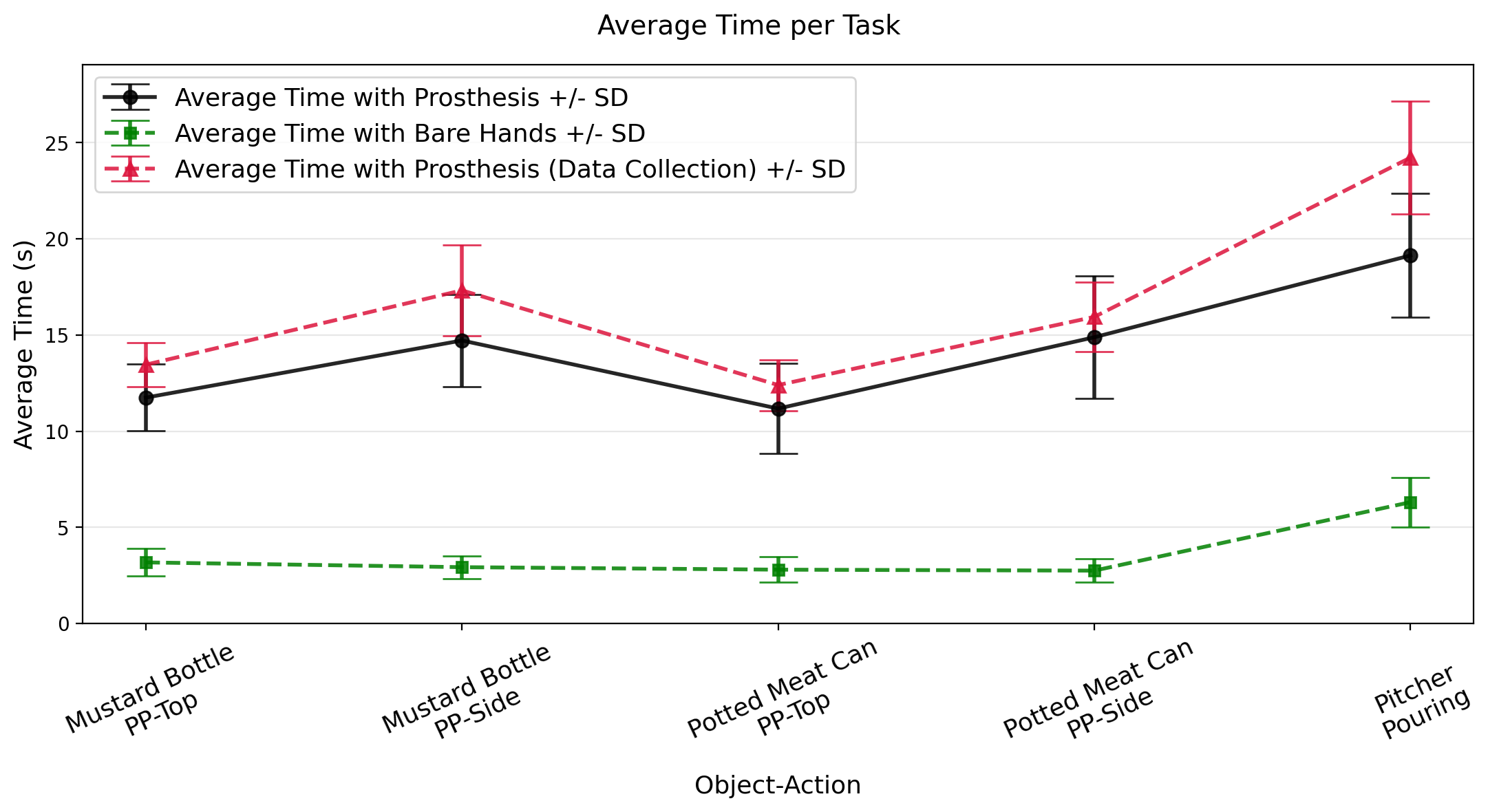}
    \caption{Mean execution time for each object-action combination in the \textit{prosthesis} and \textit{bare hands} conditions. Prosthetic execution times measured during data collection are also reported for comparison. Error bars denote standard deviation.}
    \label{fig:time_execution}
\end{figure}

Fig.~\ref{fig:time_execution} compares the mean execution time obtained in the \textit{bare hands} condition and in the \textit{prosthesis} condition.
For every object-action combination, task execution was slower when the prosthesis was used. Mean execution times in the \textit{prosthesis} condition ranged approximately from 11.3~s to 19.0~s, whereas those measured using both bare hands ranged approximately from 2.8~s to 6.4~s.

Among the pick-and-place tasks, the shortest mean execution time in the \textit{prosthesis} condition was observed for the potted meat can PP-Top (approximately 11.3~s). Side grasps required longer execution times, with mean values of approximately 14.7~s for both objects. This is because the side-grasp configurations required additional time for prosthesis alignment and grasp stabilization.

The pouring task presented the longest execution time in both conditions. Its mean duration was approximately 19~s with the prosthesis and 6.4~s with bare hands. In addition to requiring object grasping, pouring involved maintaining a stable grasp while coordinating object orientation and motion with the intact hand, making it the most temporally demanding task in the experimental set.

The error bars show substantial inter-participant variability, particularly for the PP-Side and pouring tasks. This variability may reflect differences in individual strategies, familiarity with the prosthetic device and ability to coordinate the prosthesis with the intact hand.
Nevertheless, the same general pattern was observed across all object-action combinations: prosthesis use increased execution time relative to bare hands manipulation.

The execution times measured during the user study were also compared with those obtained during the data collection session (i.e., average demonstration duration).
The two series followed a similar trend across tasks, with longer durations for PP-Side and pouring than for PP-Top grasps. However, the demonstrations collected using the two-operator setup were, on average, longer than the corresponding user-study trials. This difference is likely related to the two-operator setup used for data collection, where the operator moving the prosthesis in the workspace and the operator controlling the DoFs had to continuously coordinate their actions. Instead, during the user study with subjects, commands are generated online by the control policy, without requiring explicit coordination with a second operator. This suggests that the policy was able to operate in coordination with the participant's movements. Although this comparison does not constitute a controlled evaluation of policy-driven control against human teleoperation, it provides a promising indication that our autonomous control framework can coordinate effectively with the user's movements. 

\subsection{Questionnaire Responses}

\subsubsection{SUS Questionnaire}

\begin{table}[]
\centering
\caption{Distribution of SUS-Questionnaire responses and $\chi^2$ test results.
Arrows indicate the favorable response direction for each item.
SD: strongly disagree; D: disagree; N: neutral; A: agree; SA: strongly agree.}
\scriptsize
\setlength{\tabcolsep}{2pt}
\renewcommand{\arraystretch}{1.15}
\begin{tabularx}{\columnwidth}{
>{\raggedright\arraybackslash}m{2.7cm}
>{\centering\arraybackslash}m{0.55cm}
>{\centering\arraybackslash}m{0.55cm}
>{\centering\arraybackslash}m{0.55cm}
>{\centering\arraybackslash}m{0.55cm}
>{\centering\arraybackslash}m{0.55cm}
>{\raggedright\arraybackslash}X
}
\toprule
Item & SD & D & N & A & SA & $\chi^2$ results \\
\midrule

Q1. I think that I would like to use this system frequently ($\rightarrow$)
& 0.0\% & 15.0\% & \cellcolor{lightpink}20.0\% & \cellcolor{lightred}50.0\% & 15.0\%
& $\chi^2(4)=13.5$, $p<0.01$ \\

Q2. I found the system unnecessarily complex ($\leftarrow$)
& \cellcolor{lightred}70.0\% & \cellcolor{lightpink}25.0\% & 5.0\% & 0.0\% & 0.0\%
& $\chi^2(4)=35.5$, $p<0.001$ \\

Q3. I thought the system was easy to use ($\rightarrow$)
& 0.0\% & 0.0\% & \cellcolor{lightpink}25.0\% & \cellcolor{lightred}50.0\% & \cellcolor{lightpink}25.0\%
& $\chi^2(4)=17.5$, $p<0.01$ \\

Q4. I think that I would need the support of a technical person to use this system ($\leftarrow$)
& \cellcolor{lightpink}30.0\% & 15.0\% & \cellcolor{lightred}40.0\% & 10.0\% & 5.0\%
& $\chi^2(4)=8.5$, $p=0.07$ \\

Q5. I found the various functions in this system were well integrated ($\rightarrow$)
& 0.0\% & 0.0\% & 25.0\% & \cellcolor{lightred}45.0\% & \cellcolor{lightpink}30.0\%
& $\chi^2(4)=15.5$, $p<0.01$ \\

Q6. I thought there was too much inconsistency in this system ($\leftarrow$)
& \cellcolor{lightred}70.0\% & \cellcolor{lightpink}20.0\% & 5.0\% & 5.0\% & 0.0\%
& $\chi^2(4)=33.5$, $p<0.001$ \\

Q7. I would imagine that most people would learn to use this system very quickly ($\rightarrow$)
& 0.0\% & 0.0\% & 5.0\% & \cellcolor{lightpink}35.0\% & \cellcolor{lightred}60.0\%
& $\chi^2(4)=28.5$, $p<0.001$ \\

Q8. I found the system very cumbersome to use ($\leftarrow$)
& \cellcolor{lightpink}20.0\% & \cellcolor{lightred}65.0\% & 5.0\% & 10.0\% & 0.0\%
& $\chi^2(4)=27.5$, $p<0.001$ \\

Q9. I felt very confident using the system ($\rightarrow$)
& 0.0\% & 0.0\% & \cellcolor{lightpink}25.0\% & \cellcolor{lightred}65.0\% & 10.0\%
& $\chi^2(4)=29.5$, $p<0.001$ \\

Q10. I needed to learn a lot of things before I could get going with this system ($\leftarrow$)
& \cellcolor{lightpink}40.0\% & \cellcolor{lightred}50.0\% & 5.0\% & 5.0\% & 0.0\%
& $\chi^2(4)=21.5$, $p<0.001$ \\

\bottomrule
\end{tabularx}
\label{tab:sus_chi2}
\end{table}

The System Usability Scale (SUS) is used to assess the perceived usability of the proposed prosthetic assistance framework. Although SUS does not directly measure cognitive load, it provides an overall indication of whether the interaction was experienced as intuitive, consistent and easy to use.

The item-level response distributions indicate an overall positive perception of system usability as can be seen in Table~\ref{tab:sus_chi2}. Most participants considered the system easy to use (Q3), well integrated (Q5) and sufficiently consistent (Q6).
In addition, 95\% of participants agreed or strongly agreed that most users would learn to operate the system quickly (Q7), while 90\% disagreed or strongly disagreed that extensive preliminary learning was required (Q10).

Participants also reported a relatively high level of confidence during interaction: 75\% agreed or strongly agreed with Q9. Negative usability statements were predominantly rejected. In particular, 95\% disagreed or strongly disagreed that the system was unnecessarily complex (Q2), 90\% rejected the presence of excessive inconsistency (Q6) and 85\% disagreed or strongly disagreed that the system was cumbersome to use (Q8).

Responses to Q1 were positive but less pronounced, with 65\% of participants reporting that they would like to use the system frequently and 20\% providing a neutral response. The most heterogeneous distribution was observed for Q4, concerning the need for technical support. In this case, 40\% selected the neutral option and the response distribution did not reach the adopted significance threshold ($p=0.07$). This result may be due to the fact that the system is still a laboratory prototype which has not been engineered as a product. In any case, it indicates that, despite the favorable overall usability assessment, some participants remained uncertain about their ability to use the system completely independently.

The chi-squared ($\chi^2$) analysis showed non-uniform response distributions for all items except Q4, supporting the presence of clear response tendencies. Overall, the questionnaire results suggest that the proposed interaction was perceived as learnable, coherent and manageable after the initial familiarization phase.

\subsubsection{NASA Questionnaire}
The NASA-TLX questionnaire is used to quantify perceived workload across mental, physical, temporal, performance, effort and frustration dimensions. 
This allows us to evaluate the burden perceived in moving the prosthetic hand within the workspace and more importantly the effort required for performing such movements in harmony with the autonomous control policy for DoFs control.

\begin{figure}[h!]
    \centering
    \includegraphics[width=1\linewidth]{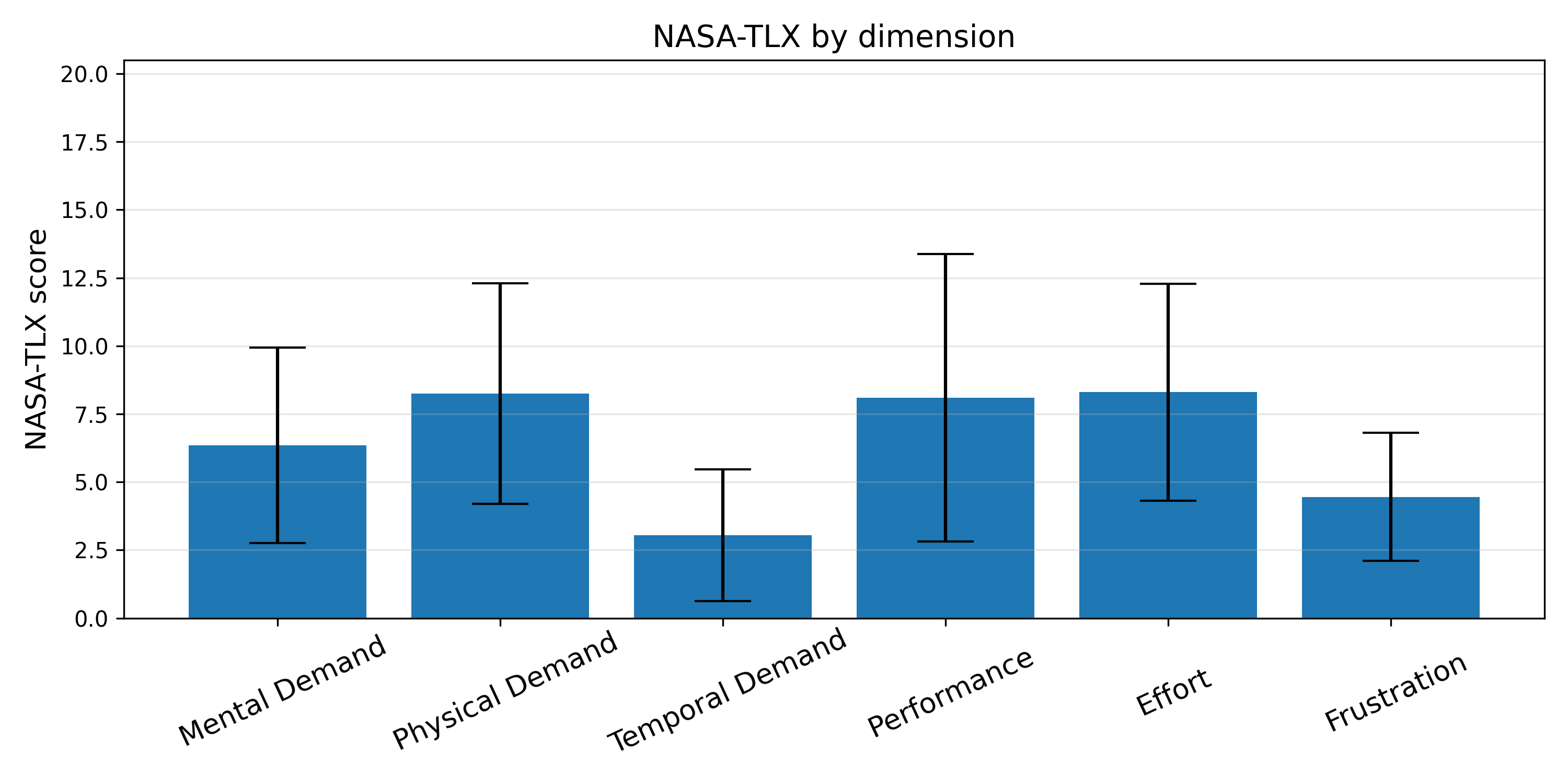}
    \caption{Mean NASA-TLX scores across 20 subjects for manipulation tasks with Hannes prosthesis. Lower scores indicate a more favorable outcome for all dimensions. Error bars denote standard deviation.}
    \label{fig:nasa_plot}
\end{figure}

Fig.~\ref{fig:nasa_plot} presents the mean scores obtained for the six NASA-TLX dimensions. Effort, physical demand and performance received the highest average ratings, with mean values of approximately 8.3, 8.25 and 8.1 respectively. Mental demand showed an intermediate mean score of approximately 6.35, while frustration and temporal demand received the lowest ratings, approximately 4.45 and 3.05, respectively.

The higher physical demand and effort scores are consistent with the requirements of operating the prosthesis using the able-bodied adapter. Although the prosthesis provided autonomous assistance, participants were still required to position the device, adapt their movements to the progression of the task, and possibly interact with their contralateral hand.

The relatively low temporal demand score suggests that participants did not generally experience strong time pressure during task execution, despite the longer durations observed in the \textit{prosthesis} condition. Similarly, frustration remained comparatively limited, indicating that the additional physical and coordination requirements did not necessarily translate into a strongly negative interaction experience.

Participants also reported a relatively positive perception of their performance, with a mean score of approximately 8.1. Note that for consistency with the other NASA-TLX dimensions, we adopted the same scale direction also for performance, with lower scores indicating a more favorable outcome.

Large standard deviations were observed, particularly for the performance, physical demand and effort dimensions. These results indicate substantial inter-participant differences in the workload experienced during prosthesis use, which may be related to individual coordination strategies, adaptation to the able-bodied adapter, training level, the weight of the prosthetic hand or previous familiarity with robotic and/or prosthetic systems.

\section{Physiological and Subjective Responses}
\label{sec:furtheranalisys}
\subsection{Exploratory Pupillometry Analysis}

Pupil-diameter variation is analyzed as a physiological correlation with cognitive and physical demand. For each participant, pupil measurements recorded during task execution are expressed relative to the baseline acquired immediately before the corresponding experimental condition (\textit{bare hands} and \textit{prosthesis} condition). This correction is used to reduce the effect of differences in absolute pupil diameter across participants and between experimental blocks.

The analysis in Table~\ref{tab:pupillometry_summary} shows a general tendency toward greater pupil dilation during prosthetic manipulation compared with bare-handed execution. This trend is consistent with the additional physical and coordination demands associated with prosthesis use. However, pupil responses varied across participants and tasks, suggesting that these demands may be experienced differently across users and task configurations. Similarly, as shown in Table~\ref{tab:pupillometry_summary}, fewer comparisons reached statistical significance when the threshold was lowered from $p<0.05$ to $p<0.001$, suggesting partial overlap in the physiological response between the two conditions. Future work could investigate whether refinements to the autonomous control policy reduce user effort, combining pupillometry with complementary measures of cognitive and physical workload. The individual pupil profiles are reported in Appendix~\ref{app:pupillometry} to provide a more detailed representation of participant-specific responses.

\begin{table}[t]
    \centering
    \caption{Cumulative summary of exploratory pupillometry comparisons across all participant--task pairs.}
    \label{tab:pupillometry_summary}
    \small
    \begin{tabular}{lccc}
    \toprule
    Outcome & $p<0.05$ & $p<0.01$ & $p<0.001$ \\
    \midrule
    \textit{Prosthesis} $>$ \textit{Bare hands} & 66/100 & 54/100 & 27/100 \\
    \textit{Prosthesis} $<$ \textit{Bare hands} & 5/100  & 4/100  & 3/100 \\
    n.s.                   & 29/100 & 42/100 & 70/100 \\
    \bottomrule
    \end{tabular}

    \vspace{1mm}
    \footnotesize{Significance columns are cumulative: comparisons with $p<0.001$ are also included in the $p<0.01$ and $p<0.05$ columns. The n.s. row reports the comparisons that did not reach significance at the corresponding confidence threshold.}
\end{table}

\subsection{Ad Hoc Questionnaire}

The ad hoc questionnaire in Table~\ref{tab:adhoc_questionnaire} is used to investigate participants' subjective experience with the prosthetic control system across five dimensions: velocity, accuracy and predictability, compensatory movements, ease of use and embodiment. The questions are specifically formulated to capture known or anticipated issues related to the control strategy and device behaviour.

\noindent \textbf{Velocity.} 
Overall, participants did not perceive the prosthesis as excessively fast, as 100\% of responses to Q1 were distributed between \textit{strongly disagree} and \textit{disagree}. However, 70\% of participants agreed that they had to perform the actions more slowly than they normally would (Q4), while 45\% reported temporary stopping of the prosthesis (Q3), this may reflect occasional uncertainty of the learned policy in ambiguous states.

\noindent \textbf{Accuracy and predictability.}
The automatic adaptation of the wrist to the object was perceived as particularly positive, with 95\% of participants agreeing that this feature was appreciable (Q5). Similarly, 80\% considered the prosthetic movements appropriate for grasping objects (Q8) and 60\% reported that the wrist generally moved as expected (Q9). However, responses were less homogeneous regarding the feeling that the prosthesis executed object grasps like the participants would have done, for which 50\% of participants expressed agreement, 30\% remained neutral and 20\% disagreed (Q7).

\noindent \textbf{Compensatory movements.}
The results concerning compensatory movements varied according to the cause considered. Most participants disagreed that compensatory movements were required either to compensate for prosthesis incorrect movement (75\%, Q10) or to allow the palm-mounted camera to detect the target object (75\%, Q11). In contrast, responses concerning the head-mounted camera were more heterogeneous, with 40\% agreeing that compensatory movements were necessary, 30\% remaining neutral and 30\% disagreeing (Q12). This indicates that the egocentric view affected participants differently. For some users, maintaining an informative head-camera view required small adjustments of the head position, for example to keep the object or the ongoing task within the camera field of view.

\noindent \textbf{Ease of use.}
The system was generally perceived as easy to control: 70\% of participants agreed that the prosthesis was easy to guide (Q13), while 85\% disagreed that the control method was frustrating (Q15). Nevertheless, the responses regarding the use of the able-bodied adapter were less conclusive, with equal proportions of neutral and positive responses, 35\% each (Q14).

\noindent \textbf{Embodiment.}
Finally, the movement of the prosthetic hand was perceived as relatively natural, with 55\% of participants agreeing or strongly agreeing with Q16. However, embodiment remained limited, as 65\% disagreed or strongly disagreed that the prosthesis felt like their own hand (Q17). Despite this, 65\% believed that the device could be used in different everyday contexts (Q18). 

In general, the participants expressed a neutral view on the need for further improvement. Most responses were neutral (60\%), while 20\% disagreed and 20\% agreed that the device could be improved. This indicates that participants did not express a clear consensus regarding the overall need for improvement.
The complete distribution of responses is reported in Appendix~\ref{app:adhoc_questionnaire}.

\begin{figure*}[b]
    \centering
    \includegraphics[width=1\linewidth]{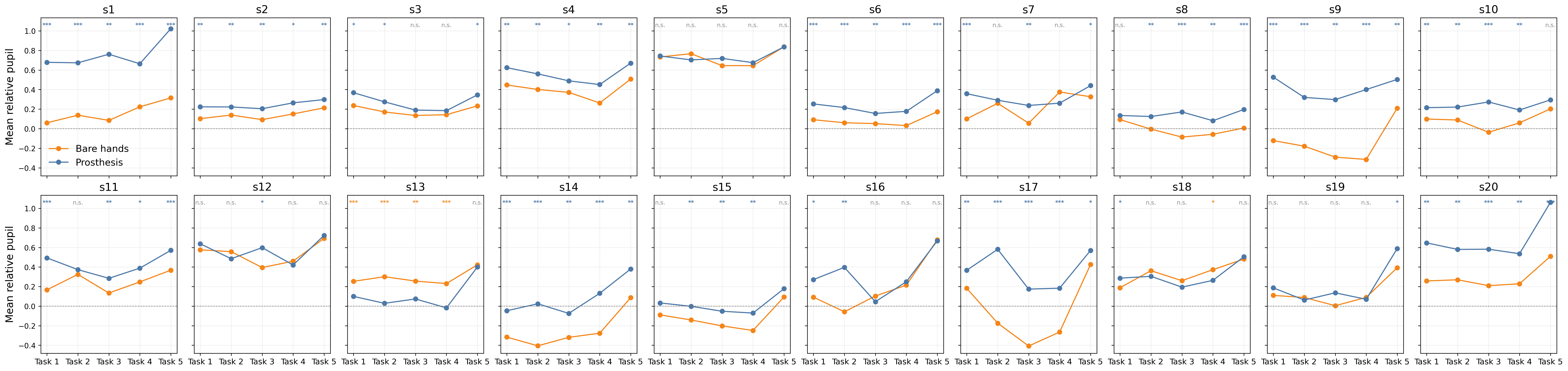}
    \caption{
    Participant-specific changes in pupil diameter relative to baseline for the \textit{bare hands} and \textit{prosthesis} conditions across the five tasks. Each subplot represents one participant. Numbers on the x-axis identify the tasks; the corresponding object--action combinations and demonstration statistics are provided in Table~\ref{tab:demo_durations}. Significance markers indicate the outcomes of exploratory within-participant comparisons, as detailed in Table~\ref{tab:pupillometry_legend}.
    }
    \label{fig:pupillometry}
\end{figure*}

\section{Discussion and Conclusion}
\label{sec:conclusions}
The results show that the proposed multimodal control policy can support prosthetic manipulation across different object-action combinations. The high SR and PSR values indicate that participants were generally able to complete the tasks or, in unsuccessful trials, to progress through most of the action sequence. Prosthesis use increased execution time and effort, as expected, but the relative trend across tasks remained similar to the bare-hands condition. This suggests that the prosthesis mainly introduced a global increase in task difficulty rather than unexpected task-specific effects. However, the need for familiarization and the uncertainty expressed by some participants in the SUS regarding independent use may indicate that the interaction is not yet fully intuitive. Future work will focus on more robust policies, a larger set of objects and actions, and evaluation with limb-difference participants.

In conclusion, this work presented an imitation learning framework for controlling the Hannes prosthetic hand during manipulation tasks. While prior vision-based prosthetic controllers have predominantly focused on grasping, our framework enables fully autonomous wrist and finger control during multi-step manipulation. The proposed multimodal diffusion policy integrates egocentric vision, eye-in-hand vision and prosthetic proprioception to generate motor commands while the user guides the hand through the workspace. Our results demonstrate that multimodal imitation learning can extend prosthetic assistance beyond isolated grasping actions to coordinated, context-aware manipulation.

\appendices

\section{Participant-Specific Pupillometry Results}
\label{app:pupillometry}

This appendix reports the baseline-corrected pupil-diameter profiles for each participant, as shown in Fig. \ref{fig:pupillometry}. Each subplot, corresponding to one participant, shows the \textit{bare hands} and \textit{prosthesis} conditions across the five object-action combination. Individual results are shown separately because the pupil response exhibited substantial inter-participant variability and could not be adequately represented by a single average plot.

Table~\ref{tab:pupillometry_legend} describes the visual elements used in the plots. The exploratory comparisons in Table~\ref{tab:pupillometry_summary} assess whether, for each participant and object-action combinations, the \textit{prosthesis} condition was associated with a greater pupil response than the \textit{bare hands} condition. A result in the direction \textit{prosthesis} $>$ \textit{bare hands} indicates greater pupil dilation during prosthesis use. We interpret this as a possible indication of increased physiological activation associated with the additional physical and coordination demands of performing the task with the prosthesis.

\begin{table}[H]
    \centering
    \caption{Interpretation of the participant-specific pupillometry plots.}
    \label{tab:pupillometry_legend}
    \begin{tabularx}{\columnwidth}{p{0.27\columnwidth}X}
    \toprule
    Element & Meaning \\
    \midrule
    Blue line & \textit{Prosthesis} condition \\
    Orange line & \textit{Bare hands} condition \\
    Y-axis & Mean baseline-corrected pupil-diameter variation \\
    X-axis & Object--action combination \\
    Blue significance marker & \textit{Prosthesis} condition significantly higher than \textit{bare hands} condition \\
    Orange significance marker & \textit{Bare hands} condition significantly higher than \textit{prosthesis} condition \\
    n.s. & No significant difference \\
    $*$, $**$, $*$$**$ & $p<0.05$, $p<0.01$, $p<0.001$ \\
    \bottomrule
    \end{tabularx}
\end{table}

\begin{table*}
\centering
\caption{Distribution of responses to the ad hoc questionnaire, grouped by macro-dimensions, with corresponding $\chi^2$ test results. Arrows indicate the favorable response direction for each item. SD: strongly disagree; D: disagree; N: neutral; A: agree; SA: strongly agree.}
\scriptsize
\setlength{\tabcolsep}{4pt}
\renewcommand{\arraystretch}{1.15}

\begin{tabularx}{0.96\textwidth}{
>{\raggedright\arraybackslash}X
>{\centering\arraybackslash}m{1.05cm}
>{\centering\arraybackslash}m{1.05cm}
>{\centering\arraybackslash}m{1.05cm}
>{\centering\arraybackslash}m{1.05cm}
>{\centering\arraybackslash}m{1.05cm}
>{\centering\arraybackslash}m{3.35cm}
}
\toprule
Item & SD & D & N & A & SA & $\chi^2$ results \\
\midrule

\rowcolor{gray!20}
\multicolumn{7}{l}{\textbf{Velocity}} \\
\addlinespace[2pt]

Q1. I had the feeling that the prosthesis was moving too fast ($\leftarrow$)
& \cellcolor{lightred}65.0\%
& \cellcolor{lightpink}35.0\%
& 0.0\%
& 0.0\%
& 0.0\%
& $\chi^2(4)=34.5$, $p<0.001$ \\

Q2. I had the feeling that the prosthesis was moving too slow ($\leftarrow$)
& \cellcolor{lightpink}20.0\%
& \cellcolor{lightred}30.0\%
& \cellcolor{lightred}30.0\%
& \cellcolor{lightpink}20.0\%
& 0.0\%
& $\chi^2(4)=6.0$, $p=0.199$ \\

Q3. I had the feeling that the prosthesis was blocked for a few seconds ($\leftarrow$)
& 15.0\%
& \cellcolor{lightpink}25.0\%
& 15.0\%
& \cellcolor{lightred}40.0\%
& 5.0\%
& $\chi^2(4)=7.0$, $p=0.136$ \\

Q4. I had to perform the actions more slowly than the speed at which I would have done it ($\leftarrow$)
& 0.0\%
& 10.0\%
& 20.0\%
& \cellcolor{lightred}40.0\%
& \cellcolor{lightpink}30.0\%
& $\chi^2(4)=10.0$, $p=0.040$ \\

\midrule

\rowcolor{gray!20}
\multicolumn{7}{l}{\textbf{Accuracy and Predictability}} \\
\addlinespace[2pt]

Q5. I appreciate the fact that the wrist automatically adapts to the object being grasped ($\rightarrow$)
& 5.0\%
& 0.0\%
& 0.0\%
& \cellcolor{lightpink}30.0\%
& \cellcolor{lightred}65.0\%
& $\chi^2(4)=31.5$, $p<0.001$ \\

Q6. I had to repeat the action several times, for example on the side-grasp, as if the prosthesis was not detecting the attempt ($\leftarrow$)
& 5.0\%
& \cellcolor{lightred}55.0\%
& 10.0\%
& \cellcolor{lightpink}25.0\%
& 5.0\%
& $\chi^2(4)=18.0$, $p<0.01$ \\

Q7. I had the feeling that the prosthesis was grasping objects just as I would have wanted to ($\rightarrow$)
& 0.0\%
& 20.0\%
& \cellcolor{lightpink}30.0\%
& \cellcolor{lightred}40.0\%
& 10.0\%
& $\chi^2(4)=10.0$, $p=0.040$ \\

Q8. I had the feeling that the movements of the prosthesis were right for grasping objects ($\rightarrow$)
& 0.0\%
& 0.0\%
& 20.0\%
& \cellcolor{lightred}55.0\%
& \cellcolor{lightpink}25.0\%
& $\chi^2(4)=20.5$, $p<0.001$ \\

Q9. The wrist always moved as I expected ($\rightarrow$)
& 0.0\%
& 10.0\%
& \cellcolor{lightpink}30.0\%
& \cellcolor{lightred}50.0\%
& 10.0\%
& $\chi^2(4)=16.0$, $p<0.01$ \\

\midrule

\rowcolor{gray!20}
\multicolumn{7}{l}{\textbf{Compensatory movements}} \\
\addlinespace[2pt]

Q10. I had to make compensatory movements with my body because the prosthesis was not moving correctly ($\leftarrow$)
& \cellcolor{lightred}40.0\%
& \cellcolor{lightpink}35.0\%
& 15.0\%
& 5.0\%
& 5.0\%
& $\chi^2(4)=11.0$, $p=0.027$ \\

Q11. I had to make compensatory movements with my body because I felt that the camera in the palm couldn't see the object I was trying to grasp ($\leftarrow$)
& \cellcolor{lightpink}35.0\%
& \cellcolor{lightred}40.0\%
& 15.0\%
& 10.0\%
& 0.0\%
& $\chi^2(4)=11.5$, $p=0.021$ \\

Q12. I had to make compensatory movements with my body because I felt that the head-mounted camera couldn't see the object I was trying to grasp ($\leftarrow$)
& 15.0\%
& 15.0\%
& \cellcolor{lightpink}30.0\%
& \cellcolor{lightred}35.0\%
& 5.0\%
& $\chi^2(4)=6.0$, $p=0.199$ \\

\midrule

\rowcolor{gray!20}
\multicolumn{7}{l}{\textbf{Ease of use}} \\
\addlinespace[2pt]

Q13. I got the feeling that the prosthesis was easy to control ($\rightarrow$)
& 0.0\%
& 10.0\%
& \cellcolor{lightpink}20.0\%
& \cellcolor{lightred}55.0\%
& 15.0\%
& $\chi^2(4)=17.5$, $p<0.01$ \\

Q14. The prosthesis was easily maneuverable with the able-bodied adapter ($\rightarrow$)
& 10.0\%
& 10.0\%
& \cellcolor{lightred}35.0\%
& \cellcolor{lightred}35.0\%
& 10.0\%
& $\chi^2(4)=7.5$, $p=0.112$ \\

Q15. For me, controlling the prosthesis using this control method was frustrating ($\leftarrow$)
& \cellcolor{lightred}60.0\%
& \cellcolor{lightpink}25.0\%
& 10.0\%
& 5.0\%
& 0.0\%
& $\chi^2(4)=23.5$, $p<0.001$ \\

\midrule

\rowcolor{gray!20}
\multicolumn{7}{l}{\textbf{Embodiment}} \\
\addlinespace[2pt]

Q16. The hand movement is quite natural ($\rightarrow$)
& 0.0\%
& 5.0\%
& \cellcolor{lightpink}40.0\%
& \cellcolor{lightred}50.0\%
& 5.0\%
& $\chi^2(4)=21.5$, $p<0.001$ \\

Q17. It felt to me as if the prosthesis was my own hand ($\rightarrow$)
& 20.0\%
& \cellcolor{lightred}45.0\%
& \cellcolor{lightpink}25.0\%
& 10.0\%
& 0.0\%
& $\chi^2(4)=11.5$, $p=0.021$ \\

Q18. I think I can use the device in various everyday situations ($\rightarrow$)
& 0.0\%
& \cellcolor{lightpink}20.0\%
& 15.0\%
& \cellcolor{lightred}45.0\%
& \cellcolor{lightpink}20.0\%
& $\chi^2(4)=10.5$, $p=0.033$ \\

\midrule

\rowcolor{gray!20}
\multicolumn{7}{l}{\textbf{Global evaluation}} \\
\addlinespace[2pt]

Q19. I believe the device could be improved in many aspects ($\leftarrow$)
& 0.0\%
& \cellcolor{lightpink}20.0\%
& \cellcolor{lightred}60.0\%
& 10.0\%
& 10.0\%
& $\chi^2(4)=22.0$, $p<0.001$ \\

\bottomrule
\end{tabularx}

\label{tab:adhoc_questionnaire}
\end{table*}


Table~\ref{tab:pupillometry_summary} summarizes the direction of these exploratory statistical comparisons across the 100 participant-task pairs (20 participants and five object-action combinations each). At the $p<0.05$ threshold, the comparisons show a significant difference between conditions associated with higher pupil dilation in the \textit{prosthesis} condition (66/100), consistently with the additional demands imposed by prosthesis use. When stricter thresholds are considered, the number of significant comparisons decreases, while the number of non-significant comparisons increases, exceeding half of the participant-task pairs at the $p<0.001$ threshold. The presence of non-significant comparisons indicates that prosthesis use does not systematically induce a detectable increase in pupil dilation for all participants and tasks. This suggests that, although the \textit{prosthesis} condition often requires greater physiological activation, the additional demand is not uniform across the experimental population. The participant-specific plots provide the corresponding visual detail and highlight the heterogeneity of pupil responses across subjects and object-action combinations, despite the presence of visible overall trends.


\section{Ad Hoc Questionnaire Responses}
\label{app:adhoc_questionnaire}

This appendix reports the questions and the responses of the ad hoc questionnaire, administered after the experimental session. The questionnaire consists of 19 items rated on a five-point Likert scale, ranging from \textit{strongly disagree} to \textit{strongly agree} and an open question about suggestions. For ease of interpretation, the items were organized into the following macro-dimensions: velocity, accuracy and predictability, compensatory movements, ease of use, embodiment and overall evaluation.

Table~\ref{tab:adhoc_questionnaire} presents the percentage of responses for each question. The most frequent response category is highlighted in dark red, while the second most frequent category is highlighted in light red. The table also reports the results of a $\chi^2$ test comparing the observed distribution with a uniform distribution across the five response categories.

Given the sample size of 20 participants and the use of five response categories, the $\chi^2$ results were interpreted as exploratory. In particular, a uniform distribution would correspond to only four participants per category, therefore, the results were primarily discussed in terms of the response percentages, while the statistical tests were used as supplementary indicators.


\section*{Acknowledgment}

This work has been carried out within the Brain and Machines Flagship of the Italian Institute of Technology.
The authors acknowledge the use of ChatGPT (OpenAI) to generate the image presented in Fig.~\ref{fig:datacollection_operators} from author-provided prompts.


\bibliographystyle{IEEEtran}
\bibliography{bibliography}


\end{document}